\documentclass[10pt,twocolumn,letterpaper]{article}

\usepackage{cvpr}              
\definecolor{cvprblue}{rgb}{0.21,0.49,0.74}
\usepackage[pagebackref,breaklinks,colorlinks,allcolors=cvprblue]{hyperref}

\def\paperID{3708} 
\def\confName{CVPR}
\def\confYear{2026}

\newcommand{\NAME}{PA-HiRes}
\usepackage{graphicx}
\usepackage{multirow}
\usepackage{amsmath}
\usepackage{bm}
\usepackage{caption}
\usepackage{array}

\title{Towards Arbitrary Motion Completing via Hierarchical Continuous Representation}

\author{Chenghao Xu$^{1}$, Guangtao Lyu$^{2}$, Qi Liu$^{2}$, Jiexi Yan$^{3}$, Muli Yang$^{4}$,  Cheng Deng$^{1}$\thanks{Corresponding author} \\
        $^{1}$ Hohai university, China, \\  
        $^{2}$ School of Electronic Engineering, Xidian University, China, \\  
        $^{3}$ School of Computer Science and Technology, Xidian University, China, \\ 
        $^{4}$ Institute for Infocomm Research (I\textsuperscript{2}R), A*STAR, Singapore 
        }

\begin{document}
\maketitle
\begin{abstract}
Physical motions are inherently continuous, and higher camera frame rates typically contribute to improved smoothness and temporal coherence. 
For the first time, we explore continuous representations of human motion sequences, featuring the ability to interpolate, inbetween, and even extrapolate any input motion sequences at arbitrary frame rates. 
To achieve this, we propose a novel \textit{parametric activation-induced hierarchical implicit representation framework}, referred to as \textbf{\textit{\NAME}}, based on Implicit Neural Representations (INRs).
Our method introduces a hierarchical temporal encoding mechanism that extracts features from motion sequences at multiple temporal scales, enabling effective capture of intricate temporal patterns. Additionally, we integrate a custom parametric activation function, powered by Fourier transformations, into the MLP-based decoder to enhance the expressiveness of the continuous representation. This parametric formulation significantly augments the model’s ability to represent complex motion behaviors with high accuracy. Extensive evaluations across several benchmark datasets demonstrate the effectiveness and robustness of our proposed approach.
\end{abstract}    
\section{Introduction}
\label{sec:intro}

3D human motion sequences hold substantial potential across a range of applications, including film production, gaming, virtual and augmented reality, and robotics, where realistic and contextually accurate human movements are critical for enhancing interactivity and immersion. At present, manually crafted and synthesized human motion data are typically stored and represented as sequences with a fixed frame rate, where the trade-off between complexity and fidelity is regulated by the chosen frame rate. However, the real-world visual experience is inherently continuous, and higher frame rates generally lead to improved smoothness and coherence. Fixed-frame-rate motion sequences may thus compromise fidelity.

To address this limitation, we intend to study a physics-inspired continuous representation for human motions. By modeling a motion sequence as a function defined in a continuous latent space, we enable the restoration and generation of motion sequences at arbitrary frame rates as required.

Implicit Neural Representations (INRs)~\cite{sitzmann2020implicit,strumpler2022implicit,dupont2021coin} offer an intuitive approach for modeling human motion sequences in a continuous manner, as they have been widely adopted for representing a variety of signals ( such as images~\cite{chen2021learning,saragadam2023wire}, videos~\cite{chen2021nerv,chen2022videoinr}, and 3D scenes~\cite{mildenhall2020nerf} ) through continuous data mappings. By modeling images and videos as continuous functions parameterized by neural networks, INRs enable reconstruction at arbitrary spatial resolutions or temporal frames without dependence on fixed discrete grids. For instance, LIIF~\cite{chen2021learning} achieves continuous image upsampling through local implicit functions, while NeRV~\cite{chen2021nerv} and HNeRV~\cite{chen2023hnerv} represent videos by mapping temporal indices to high-resolution frames, implicitly capturing temporal dependencies. Similarly, VideoINR~\cite{chen2022videoinr} encodes videos as continuous space-time functions, facilitating both spatial and temporal super-resolution at arbitrary scales. 

\begin{figure*}[t]
    \centering
\includegraphics[width = 1.0\textwidth]{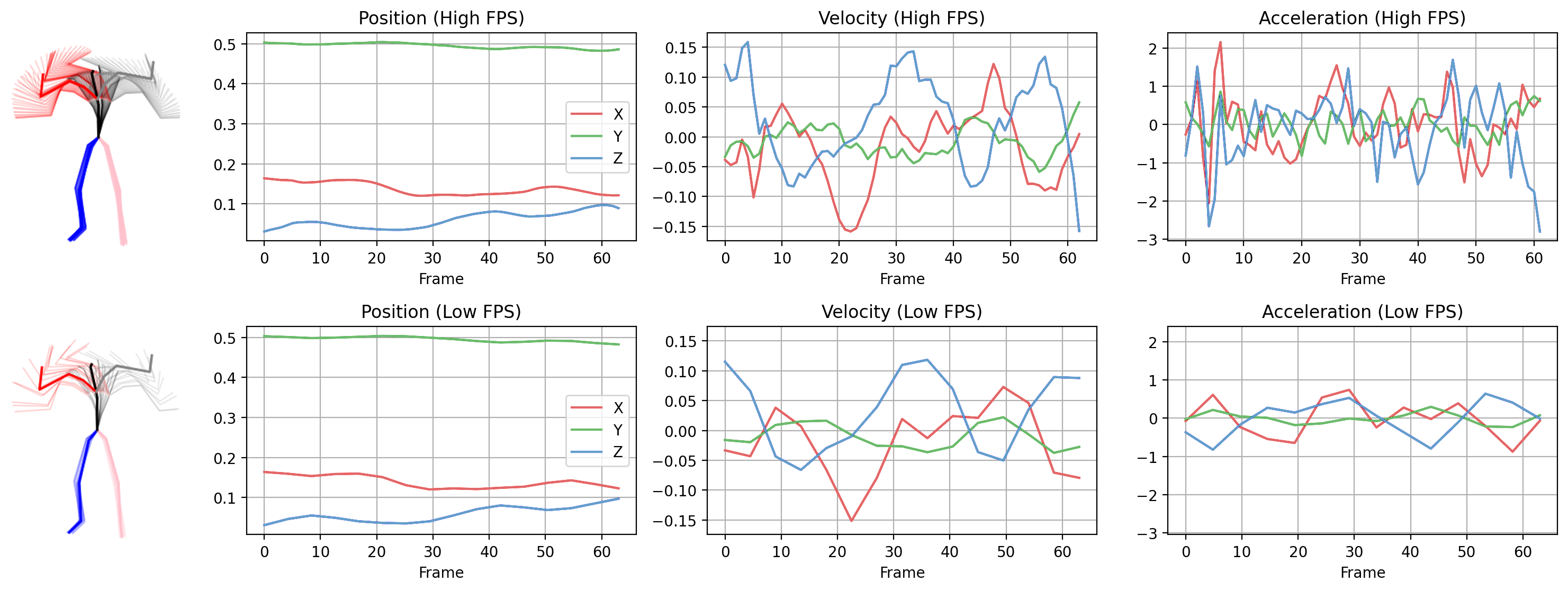}
    \caption{\textbf{Illustration of complex physical characteristics of human motion sequence across different frame rates.} Human motion sequences inherently exhibit rich and intricate physical properties, such as velocity and acceleration, which vary significantly with changes in frame rate. While the positional data of motions sampled at different FPS (frames per second) may appear visually similar, their underlying physical dynamics, particularly temporal derivatives like velocity and acceleration, can differ substantially. }
    \label{fig:intro}
\end{figure*}

However, existing INR-based methods are predominantly tailored for image or video data, and their direct application to human motion sequences is limited due to fundamental differences in the structural and temporal characteristics of motion data compared to images and videos. Unlike images or videos, human motion sequences exhibit inherent physical properties that are crucial for accurate representation. As demonstrated in Figure~\ref{fig:intro}, while positional information remains relatively consistent across motion sequences with different frame rates, significant discrepancies are observed in velocity and acceleration. Therefore, an effective continuous implicit representation must account for these differences to achieve smoother and higher-fidelity motion reconstruction.

In this paper, we introduce a novel \textit{parametric activation-induced hierarchical implicit representation framework}, termed \textbf{\textit{\NAME}}, designed to effectively model continuous human motion sequences at arbitrary frame rates. The proposed \textbf{\textit{\NAME}} consists of two primary components: the Multi-Scale Temporal Encoding (MSTE) module and the Parametric Activation Induced Decoding (PAID) module. Recognizing the limitations of using the nearest motion feature as a reference, the MSTE module employs a hierarchy of temporal encoders to extract features from human motion clips at multiple temporal scales, thereby capturing rich temporal dynamics. These multi-scale features are then integrated in the PAID module to produce fused reference vectors, which, along with temporal coordinates, are input to a decoder equipped with a carefully designed parametric activation function for constructing the continuous implicit representation. To further improve the expressiveness of the continuous implicit representations, we incorporate parametric activation functions into the MLP architecture, significantly enhancing the capacity of the model to represent complex motion characteristics with high fidelity. 


In summary, our main contributions include:
\begin{itemize}
    \item We investigate continuous implicit representations for high-fidelity human motion sequences across arbitrary frame rates.

    \item We propose a novel parametric activation-induced hierarchical implicit representation framework, termed \textbf{\textit{\NAME}}, which hierarchically captures rich temporal dynamics to accurately model the complex physical characteristics inherent in human motion sequences.

    \item Comprehensive experiments on public datasets demonstrate that our method achieves state-of-the-art performance, confirming its effectiveness.
\end{itemize}
\section{Related Work}
\label{sec:formatting}

\subsection{Motion Generation and Editing}
Text-driven human motion generation~\cite{dai2024motionlcm,lu2023humantomato,guo2024momask,tevethuman} aims to synthesize realistic human motions conditioned on textual descriptions. Two dominant paradigms have emerged: GPT-like auto-regressive models~\cite{zhang2023generating,lu2023humantomato,guo2024momask}, which generate motion sequentially based on a single conditional embedding, and diffusion-based approaches~\cite{tevethuman,chen2023executing}, often combined with transformer architectures. Despite the rapid progress in these methods, the interpretability of attention mechanisms—especially the fine-grained correspondence between text and motion—remains underexplored.

Motion Editing aims to modify motion sequences to meet user-specified requirements. Previous works~\cite{dai2024motionlcm,kim2023flame} attempt to edit motions in a controlled manner, such as through motion inbetweening or joint-level manipulation. Other methods focus on altering stylistic attributes of motion~\cite{jang2022motion,raab2023modi}. 
Recent studies have explored more flexible editing strategies. Raab et al.~\cite{raab2023modi} manipulate self-attention queries to guide motion following, while Goel et al.~\cite{goel2024iterative} propose instruction-based motion editing. Despite these advances, the fine-grained correspondence between text and motion in cross-attention mechanisms remains insufficiently understood.

\subsection{Activation Functions in INR}
INRs have advanced in representing various signals, including images and 3D scenes, with applications in SDFs, audio signals, and data compression.
Due to the continuous property of INR, the development of neural networks has been significantly influenced by advancements in activation functions. 
Early non-periodic functions like Sigmoid suffered from vanishing gradient issues in deep networks, which were later addressed by unbounded functions such as ReLU~\cite{nair2010rectified} and its variants~\cite{hendrycks2016gaussian,elfwing2018sigmoid}. Adaptive functions like SinLU~\cite{paul2022sinlu} and Swish~\cite{ramachandran2017searching} introduced trainable parameters to better adapt to data non-linearity. However, the spectral bias in ReLU-based networks, as highlighted by Rahaman et al.~\cite{rahaman2019spectral}, led to a preference for low-frequency signals, limiting their ability to capture fine details.
To address this, periodic activation functions emerged as promising solutions for INRs, enabling the learning of high-frequency details. Early challenges in training networks with periodic activations~\cite{lapedes1987nonlinear} were eventually overcome, leading to successful applications in complex data representation~\cite{sitzmann2020implicit,mehta2021modulated}. Recently, the Kolmogorov-Arnold Network (KAN)~\cite{liu2024kan,liu2024kan2} has emerged as a promising architecture in the realm of INRs. KAN leverages Kolmogorov-Arnold representation frameworks to improve the modeling and reconstruction of complex signals, demonstrating notable performance in various INR tasks.

\section{Method}

\subsection{Preliminaries}

\paragraph{Implicit Neural Representation for Human Motion Sequences.} To employ INRs for continuous modeling of human motion sequences, we adopt a shared decoding function \( f_{\theta} \), parameterized by \( \theta \), implemented as a MLP and jointly optimized across all motion sequences. Specifically, given a motion sequence $ \mathbf{M}_i \in \mathbb{R}^{T\times D}$, the formula of $f_{\theta}$ is represented as follows:
\begin{equation}
    \bm{m}_t = f_{\theta}(t, \bm{r}) \in \mathbb{R}^D,
\end{equation}
where $\bm{r}$ is a reference vector, $t \in \mathcal{T} $ is the temporal coordinate in the continuous motion sequence, and $\bm{m}_t \in \mathcal{M}$ is the predicted motion feature corresponding to the temporal coordinate $t$. Given the defined function \( f_{\theta}: \mathcal{T} \rightarrow \mathcal{M} \), each feature vector \( \bm{m}_t \) can be interpreted as a functional representation that maps temporal coordinates to corresponding motion features. 

To effectively predict the motion feature \( \bm{m}_t \), it is necessary to leverage available reference information. The most straightforward approach is to use the motion feature corresponding to the temporal coordinate closest to \( t \) as the reference input $\bm{r}$. However, this coarse approach relies on overly limited reference information and overlooks important characteristics of the motion sequence, thereby failing to achieve satisfactory performance.

\paragraph{Skeleton-aware Human Motion Embeddings.}
Previous works in text-driven motion generation~\cite{guo2024momask,lu2023humantomato}, text-motion retrieval~\cite{petrovich2023tmr,lyu2025towards}, and text-driven motion editing~\cite{athanasiou2024motionfix,goel2024iterative} have predominantly relied on global feature representations to model human motion, often treating poses as single vectors and neglecting the complex interdependencies between skeletal joints and temporal frames. 
In contrast, we use a skeleton-aware representation that explicitly captures the spatial relationships among joints and their temporal evolution.
This approach enables a more structured and detailed modeling of motion data, preserving anatomical constraints and enhancing the fidelity of output motions.

\begin{figure*}[t]
   \centering
\includegraphics[width = 0.98\textwidth]{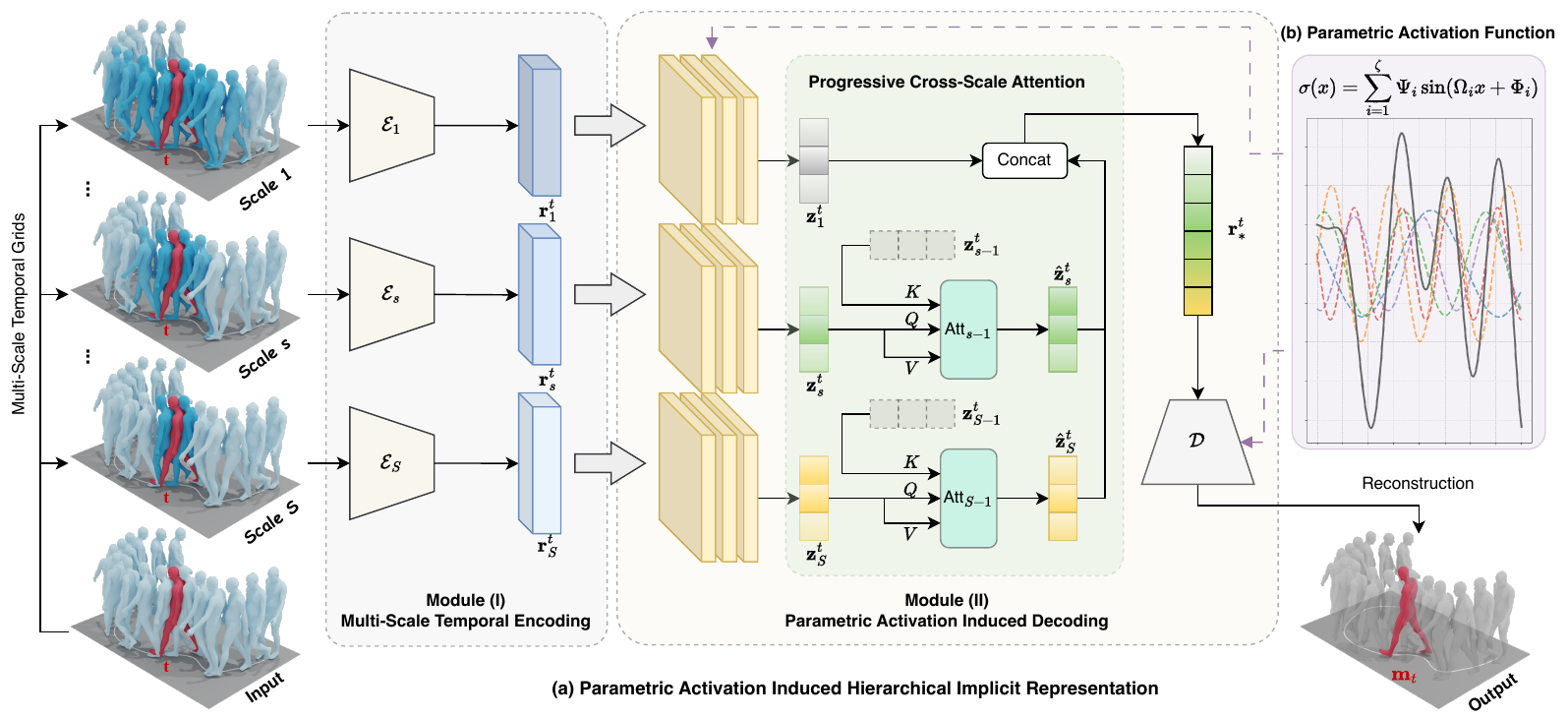}
  \caption{(a) The overall framework of our \NAME. (b) The simple illustration and visualization of the proposed parametric activation function.}
   \label{fig:framework}
\end{figure*}

\subsection{Parametric Activation Induced Hierarchical Implicit Representation}
Existing general-purpose or image/video-specific INR-based continuous representation methods fail to capture the essential characteristics of human motion sequences. Given the complex and nuanced variations in multiple physical quantities across different frame rates, as illustrated in Figure~\ref{fig:intro}, we propose a novel physics-informed implicit representation method, termed \textbf{\textit{\NAME}}, which effectively models continuous human motion sequences at arbitrary frame rates.

\paragraph{Overall.} As illustrated in Figure~\ref{fig:framework}, our proposed \textbf{\textit{\NAME}} framework comprises two key components: the Multi-Scale Temporal Encoding (MSTE) module and the Parametric Activation Induced Decoding (PAID) module. The MSTE module employs a set of hierarchical encoders to extract temporal features from human motion clips at multiple scales, effectively capturing temporal dynamics. Subsequently, the PAID module integrates these multi-scale features to generate fused reference feature vectors, which are combined with temporal coordinates and passed into a decoder equipped with a specially designed parametric activation function to construct the implicit representation of the human motion sequence.

\paragraph{Multi-scale Temporal Encoding.} Using the motion feature associated with the temporal coordinate nearest to $t$ as the reference input $\bm{r}$ proves insufficient, as it provides overly limited contextual information. To address this limitation and effectively capture the complex and nuanced variations across multiple physical quantities in human motion sequences, we propose the use of a set of hierarchical encoders $\lbrace \mathcal{E}_s \rbrace_{s=1}^S$ that operate on multi-scale local clips. This design enables the extraction of rich temporal features that better represent the intricate dynamics inherent in human motion.

We utilize multi-scale temporal grids to derive local reference vectors. Specifically, given an input temporal coordinate $t$, we first extract a set of local motion clips $\{ \mathcal{C}_1^t, \mathcal{C}_2^t, \cdots, \mathcal{C}_S^t \}$ based on a series of multi-scale temporal grids centered at $t$. Subsequently, we derive a reference feature vector corresponding to each of these multi-scale temporal clips as follows:
\begin{equation}
    \bm{r}^t_s = \mathcal{E}_s (\mathcal{C}_s^t,t),
\end{equation}
where a local clip centered around $t$ at scale level $s$ consists of $N_s$ feature vectors of motion frames. 

\paragraph{Parametric Activation Induced MLP.} To enhance the representational capacity of the extracted multi-scale reference vectors, we utilize a set of parametric activation-induced MLPs, denoted as $\{ \mathcal{F}_s \}_{s=1}^S$, to embed these reference vectors into a series of informative latent codes as follows:
\begin{equation}
    \bm{z}_s^t = \mathcal{F}_s(\bm{r}_s^t),
\end{equation}
where $\bm{z}_s^t$ is the latent code corresponding to $\bm{r}^t_s$.

In conventional MLPs, commonly used activation functions such as ReLU often struggle to capture high-frequency components, thereby limiting the model’s ability to accurately represent and reconstruct the complex physical variations inherent in human motion. To address this issue, we propose replacing traditional non-parametric activation functions with learnable parametric counterparts.
Inspired by KAN~\cite{liu2024kan,liu2024kan2}, we introduce a novel \textit{Fourier-based parametric activation function}, which is parametrized in the form of a Fourier series as follows:
\begin{equation}
    \sigma(\bm{x}) = \sum_{i=1}^{\zeta} \bm{\Psi}_i \sin ( \bm{\Omega}_i\bm{x} + \bm{\Phi}_i ), 
\end{equation}
where $\bm{\Psi}_i$, $\bm{\Omega}_i$, and $\bm{\Phi}_i$ represent the amplitude, frequency, and phase parameters, respectively. These parameters are learned dynamically during training, allowing the MLPs to adapt their activation functions to the specific characteristics present at different temporal scales. This adaptability enables the resulting latent codes to achieve a compact and flexible representational capacity for capturing the complex patterns inherent in human motion.

To facilitate more efficient optimization, we propose an initialization strategy specifically designed for networks employing our parametric activation function. This strategy aims to enhance both the performance and stability of the network during training. Detailed descriptions of the initialization procedure are provided in the supplementary material.

\paragraph{Progressive Cross-Scale Attention.} To effectively integrate multi-scale local information embedded within the hierarchical reference vectors, we adopt a top-down progressive cross-attention mechanism to fuse features across different temporal resolutions. Specifically, we utilize a sequence of $S - 1$ cross-attention blocks $\{ \text{Att}_s \}_{s=1}^{S-1}$ to iteratively refine the latent codes as follows:

\begin{equation}
    \begin{aligned}
        \tilde{\bm{z}}^t_1 &= \bm{z}^t_1 \\
        \tilde{\bm{z}}^t_2 &= \text{Att}_1(\bm{z}^t_1, \bm{z}^t_2, \bm{z}^t_2), \\
        & \cdots \\
        \tilde{\bm{z}}^t_S &= \text{Att}_{S-1}(\bm{z}^t_{S-1}, \bm{z}^t_S, \bm{z}^t_S).
\end{aligned}
\end{equation}

\begin{table*}[t]
    \centering
    \caption{Quantitative comparison for integer-scale interpolation with PSNR(dB)($\uparrow$) and SSIM($\uparrow$). Comparison with other INR-based Methods on different datasets and scales.}
    \setlength{\tabcolsep}{1.5mm}
    \renewcommand\arraystretch{0.8}
    \label{tab:results}
    \begin{tabular}{l l c c c c c c c c}
        \toprule
        \multirow{2}{*}{Datasets} & \multirow{2}{*}{Methods} & \multicolumn{2}{c}{Scale $\times$2} & \multicolumn{2}{c}{Scale $\times$3} & \multicolumn{2}{c}{Scale $\times$4} & \multicolumn{2}{c}{Scale $\times$5} \\
        \cmidrule(lr){3-4} \cmidrule(lr){5-6} \cmidrule(lr){7-8} \cmidrule(lr){9-10}
        & & PSNR & SSIM & PSNR & SSIM & PSNR & SSIM & PSNR & SSIM \\
        \midrule
        \multirow{4}{*}{HumanML3d} & Meta-SR~\cite{hu2019meta} & 31.940 & 0.980 & 27.071 & 0.944 & 25.855 & 0.931 & 23.915 & 0.910 \\
        & LIIF~\cite{chen2021learning} & 33.971 & 0.986 & 29.062 & 0.967 & 26.681 & 0.939 & 25.108 & 0.924 \\
        & ALIIF~\cite{li2022adaptive} & 34.154 & 0.987 & 29.235 & 0.970 & 27.461 & 0.942 & 25.844 & 0.926 \\
        & LMF~\cite{he2024latent} & 34.734 & 0.989 & 30.686 & 0.975 & 28.457 & 0.951 & 26.683 & 0.931 \\
        & Ours & \textbf{36.982} & \textbf{0.994} & \textbf{33.892} & \textbf{0.983} & \textbf{31.030} & \textbf{0.975} & \textbf{28.631} & \textbf{0.947} \\
        \midrule
        \multirow{4}{*}{LaFAN1} & Meta-SR & 30.624 & 0.974 & 26.012 & 0.940 & 24.462 & 0.923 & 22.234 & 0.899 \\
        & LIIF & 32.031 & 0.979 & 28.076 & 0.959 & 25.001 & 0.928 & 23.597 & 0.901 \\
        & ALIIF & 32.174 & 0.980 & 28.387 & 0.960 & 25.471 & 0.929 & 23.718 & 0.902 \\
        & LMF & 33.153 & 0.982 & 29.423 & 0.970 & 26.612 & 0.934 & 25.305 & 0.925 \\
        & Ours & \textbf{35.694} & \textbf{0.988} & \textbf{32.499} & \textbf{0.981} & \textbf{29.163} & \textbf{0.950} & \textbf{27.830} & \textbf{0.936} \\
        \midrule
        \multirow{4}{*}{CMU} & Meta-SR & 32.251 & 0.982 & 28.448 & 0.958 & 26.162 & 0.950 & 24.373 & 0.928 \\
        & LIIF & 34.648 & 0.987 & 29.063 & 0.970 & 27.244 & 0.965 & 26.023 & 0.948 \\
        & ALIIF & 34.786 & 0.988 & 29.125 & 0.971 & 27.736 & 0.967 & 26.299 & 0.948 \\
        & LMF &  35.991 & 0.991 & 31.387 & 0.978 & 28.881 & 0.971 & 27.754 & 0.965 \\
        & Ours & \textbf{37.542} & \textbf{0.995} & \textbf{34.930} & \textbf{0.987} & \textbf{31.229} & \textbf{0.979} & \textbf{30.366} & \textbf{0.972} \\
        \bottomrule
    \end{tabular}
\end{table*}

Once the latent codes have been enhanced through cross-scale attention, we concatenate the outputs from all temporal levels to construct the final fused reference feature vector as follows:

\begin{equation}
   \bm{r}^t_{*} = \text{Concat}(\tilde{\bm{z}}^t_1, \tilde{\bm{z}}^t_2, \cdots, \tilde{\bm{z}}^t_S). 
\end{equation}

\paragraph{Fused Reference oriented Decoding.} The final motion feature $\bm{m}_t$ corresponding to the $t$-th frame is predicted using a decoder $\mathcal{D}$, implemented as a multilayer perceptron (MLP), as follows:
\begin{equation}
        \bm{m}_t = \mathcal{D}(\bm{r}^t_{*}, t).
\end{equation}
To enhance the representational capacity of the learned continuous implicit representations, we further incorporate the parametric activation function introduced above within the decoder architecture.

\subsection{Continuous Human Motion Modeling} During training, a collection of human motion sequences is utilized as the training set, with the objective of learning a continuous implicit representation through our proposed \textit{\NAME} framework that generalizes to unseen motion sequences. Specifically, given a training human motion sequence, we query the predicted motion feature $\bm{m}_t$ at a temporal coordinate $t$ from the learned continuous implicit representation. The prediction is then compared with the ground-truth motion feature $\hat{\bm{m}}_t$, and the mean squared error (MSE) loss is computed as follows:
\begin{equation}
    \mathcal{L}_{\text{MSE}} = \sum_{t=1}^T ||\bm{m}_t -\hat{\bm{m}}_t||_2^2.
\end{equation}
To further improve the physical plausibility of the generated motion outputs, we incorporate a velocity consistency constraint defined as follows:
\begin{equation}
\mathcal{L}_{V} = \sum_{t=1}^{T-1} || \bm{v}_t - \hat{\bm{v}}_t ||_2^2,
\end{equation}
where $\bm{v}_t = \bm{m}_{t+1} - \bm{m}_t$ and $\hat{\bm{v}}_t = \hat{\bm{m}}_{t+1} - \hat{\bm{m}}_t$ represent the predicted and ground-truth velocities, respectively. Overall, the loss function is summarized as follows:
\begin{equation}
    \mathcal{L} = \mathcal{L}_{\text{MSE}} + \lambda \mathcal{L}_{V},
\end{equation}
where $\lambda$ is a trade-off hyperparameter.


\section{Experiments}
\label{sec:exp}


\subsection{Implementation Details}

\textbf{Experimental Implementation.} 
To simulate a continuous magnification process, the downsampling factor is randomly sampled from a uniform distribution, $U(1, 4)$, enabling the model to adapt to different degrees of motion degradation.
We train our models on an NVIDIA A100 with a batch size of 256. We use Adam optimizer~\cite{ADAMopt} with a learning rate as 0.0001. The models are trained for 1000 epochs, and the learning rate decays by a factor of 0.5 every 200 epochs.
The decoding function $\mathcal{D}$ is a 5-layer MLP and hidden dimensions of 256.
We set $S$, $\zeta$ and $\lambda$ to 3, 16 and 0.5.

\begin{table*}[t]
    \centering
    \caption{Quantitative comparison for noninteger-scale interpolation with PSNR(dB)($\uparrow$) and SSIM($\uparrow$). Comparison with other INR-based Methods on different datasets and scales.}
    \setlength{\tabcolsep}{1.5mm}
    \renewcommand\arraystretch{0.8}
    \label{tab:results_noninteger}
    \begin{tabular}{l l c c c c c c c c}
        \toprule
        \multirow{2}{*}{Datasets} & \multirow{2}{*}{Methods} & \multicolumn{2}{c}{Scale $\times$1.2} & \multicolumn{2}{c}{Scale $\times$2.4} & \multicolumn{2}{c}{Scale $\times$3.6} & \multicolumn{2}{c}{Scale $\times$4.8} \\
        \cmidrule(lr){3-4} \cmidrule(lr){5-6} \cmidrule(lr){7-8} \cmidrule(lr){9-10}
        & & PSNR & SSIM & PSNR & SSIM & PSNR & SSIM & PSNR & SSIM \\
        \midrule
        \multirow{4}{*}{HumanML3d} & Meta-SR & 32.388 & 0.981 & 27.630 & 0.951 & 25.929 & 0.931 & 24.035 & 0.911 \\
        & LIIF & 33.648 & 0.984 & 28.833 & 0.970 & 26.938 & 0.941 & 25.354 & 0.926 \\
        & ALIIF & 33.791 & 0.986 & 29.347 & 0.972 & 27.938 & 0.944 & 25.811 & 0.928 \\
        & LMF & 34.934 & 0.989 & 31.186 & 0.977 & 28.703 & 0.953 & 27.438 & 0.938 \\
        & Ours & \textbf{36.645} & \textbf{0.992} & \textbf{34.148} & \textbf{0.985} & \textbf{31.492} & \textbf{0.967} & \textbf{29.572} & \textbf{0.959} \\
        \midrule
        \multirow{4}{*}{LaFAN1} & Meta-SR & 31.771 & 0.980 & 26.868 & 0.943 & 24.879 & 0.926 & 22.514 & 0.901 \\
        & LIIF & 32.844 & 0.981 & 28.826 & 0.961 & 25.382 & 0.930 & 23.608 & 0.906 \\
        & ALIIF & 32.903 & 0.981 & 28.990 & 0.962 & 25.578 & 0.930 & 23.893 & 0.907 \\
        & LMF & 35.215 & 0.985 & 31.192 & 0.973 & 26.964 & 0.936 & 25.862 & 0.929 \\
        & Ours & \textbf{36.002} & \textbf{0.988} & \textbf{33.313} & \textbf{0.984} & \textbf{29.751} & \textbf{0.954} & \textbf{28.451} & \textbf{0.938} \\
        \midrule
        \multirow{4}{*}{CMU} & Meta-SR & 34.257 & 0.987 & 31.725 & 0.977 & 27.288 & 0.952 & 24.532 & 0.929 \\
        & LIIF & 34.863 & 0.988 & 32.883 & 0.979 & 28.583 & 0.968 & 26.237 & 0.949 \\
        & ALIIF & 35.015 & 0.990 & 32.926 & 0.980 & 28.863 & 0.969 & 26.754 & 0.950 \\
        & LMF & 36.157 & 0.992 & 33.243 & 0.986 & 29.036 & 0.974 & 28.232 & 0.969 \\
        & Ours & \textbf{38.428} & \textbf{0.996} & \textbf{35.977} & \textbf{0.988} & \textbf{32.818} & \textbf{0.982} & \textbf{30.702} & \textbf{0.975} \\
        \bottomrule
    \end{tabular}
\end{table*}

\textbf{Datasets.}
We conduct experiments on three widely used motion datasets: HumanML3D~\cite{guo2022generating}, LaFAN1~\cite{harvey2020robust}, and CMU Mocap\footnote{Dataset available at: \url{http://mocap.cs.cmu.edu/}}. The motion sequences in HumanML3D are captured at 20 frames per second (fps), LaFAN1 at 30 fps, and CMU Mocap at 120 fps. For HumanML3D and CMU Mocap, we use the standard dataset splits for evaluation. For LaFAN1, we follow the protocol where data from subjects 1 to 4 is used for training, and subject 5 is used for validation.



\begin{table}
  \caption{Results on LaFAN1 dataset. A lower score is better.}
  \label{inbetween}
  \centering
\setlength{\tabcolsep}{0.3mm}
\renewcommand\arraystretch{1.0}
\begin{tabular}{l|ccccccccc}
\toprule
Metrics    & \multicolumn{3}{c}{L2P($\downarrow$)} & \multicolumn{3}{c}{L2Q($\downarrow$)} & \multicolumn{3}{c}{NPSS($\%$)($\downarrow$)} \\
\cmidrule(lr){2-4} \cmidrule(lr){5-7} \cmidrule(lr){8-10} 
Lengths   & 5      & 15     & 30    & 5      & 15     & 30    & 5      & 15     & 30     \\
\midrule
MC-Trans  & 0.23   & 0.74   & 1.37  & 0.17   & 0.42   & 0.69  & 0.19   & 2.91   & 14.30  \\
NeMF       & 0.25   & 0.69   & 1.30  & 0.18   & 0.40   & 0.62  & 0.22   & 2.75   & 13.77  \\
ERD-QV    & 0.23   & 0.65   & 1.28  & 0.17   & 0.42   & 0.69  & 0.20   & 2.58   & 13.28  \\
$\Delta$-Interp  & 0.13   & 0.47   & 1.00  & 0.11   & 0.32   & 0.57  & 0.14   & 2.17   & 12.17  \\
TS-Former & 0.10   & 0.39   & 0.89  & 0.10   & 0.28   & 0.54  & 0.11   & 1.88   & 11.24  \\
Ours      & \textbf{0.09}   & \textbf{0.36}   & \textbf{0.85}  & \textbf{0.09}   & \textbf{0.27}   & \textbf{0.52}  & \textbf{0.10}   & \textbf{1.80}   & \textbf{10.55}  \\
\bottomrule
\end{tabular}
\end{table}

\subsection{Motion Sequence Interpolation}

\paragraph{Evaluation Metrics.}
To quantitatively evaluate the quality of the reconstructed or generated images, we adopt two widely used image fidelity metrics: Peak Signal-to-Noise Ratio (PSNR) and Structural Similarity Index (SSIM). PSNR measures the pixel-wise reconstruction accuracy by comparing the maximum possible signal power to the power of the residual noise, providing a general indication of distortion levels. A higher PSNR value indicates a better reconstruction with less noise or artifacts. On the other hand, SSIM evaluates the structural information preservation between two motions, focusing on luminance, contrast, and structural consistency. It provides a more perceptually meaningful assessment of visual quality, especially in capturing fine details and natural motion structures. 

\begin{table}[t]
\centering
\caption{Ablation study for PAID.}
\renewcommand\arraystretch{0.9}
\setlength{\tabcolsep}{3mm}
\begin{tabular}{c|cc|cc}
\toprule
Metrics & \multicolumn{2}{c|}{PSNR} & \multicolumn{2}{c}{SSIM} \\
\midrule
Share & $\times$ & $\checkmark$ & $\times$ & $\checkmark$ \\
\midrule
1  & 28.10 & 27.67 & 0.949 & 0.939 \\
2  & 29.12 & 28.04 & 0.956 & 0.945 \\
4  & 29.75 & 28.53 & 0.968 & 0.951 \\
8  & 30.29 & 29.03 & 0.971 & 0.955 \\
16 & 30.53 & 29.33 & 0.972 & 0.957 \\
32 & 30.66 & 29.51 & 0.973 & 0.958 \\
\bottomrule
\end{tabular}
\label{tab:zeta}
\end{table}

\begin{figure*}[t]
   \centering
\includegraphics[width = 0.85\textwidth]{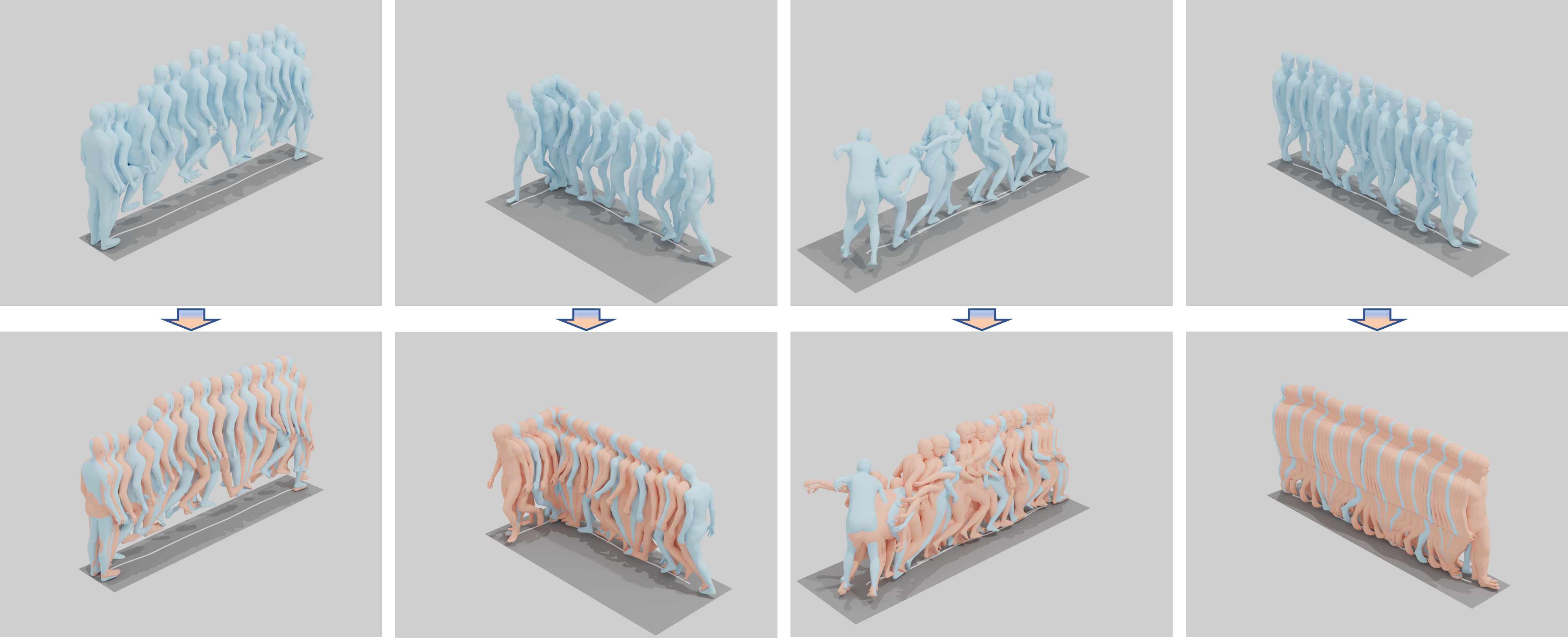}
  \caption{Visualization results under $\times$2, $\times$3, $\times$4, and $\times$5 interpolation. The blue motions represent the known input, while the red motions indicate the results generated by our \NAME.}
   \label{fig:results}
\end{figure*}

\textbf{Quantitative Results.} To evaluate the effectiveness of our method, we conduct motion sequence interpolation experiments on three benchmarks including HumanML3D~\cite{guo2022generating}, LaFAN1~\cite{harvey2020robust}, and CMU Mocap. We compare our method with general-purpose and image-specific INR-based methods Meta-SR, LIIF, ALIIF, and LMF. Tables~\ref{fig:results} and~\ref{tab:results_noninteger} report the performance of our method on integer-scale and non-integer-scale interpolation tasks, respectively. Our approach demonstrates consistently competitive results across all three datasets and scaling factors, with particularly strong performance on non-integer scaling factors, attributed to the flexibility of the proposed representation.


\begin{figure*}[t]
\centering
\begin{minipage}[c]{0.25\textwidth}
  \centering
  \includegraphics[width=\linewidth]{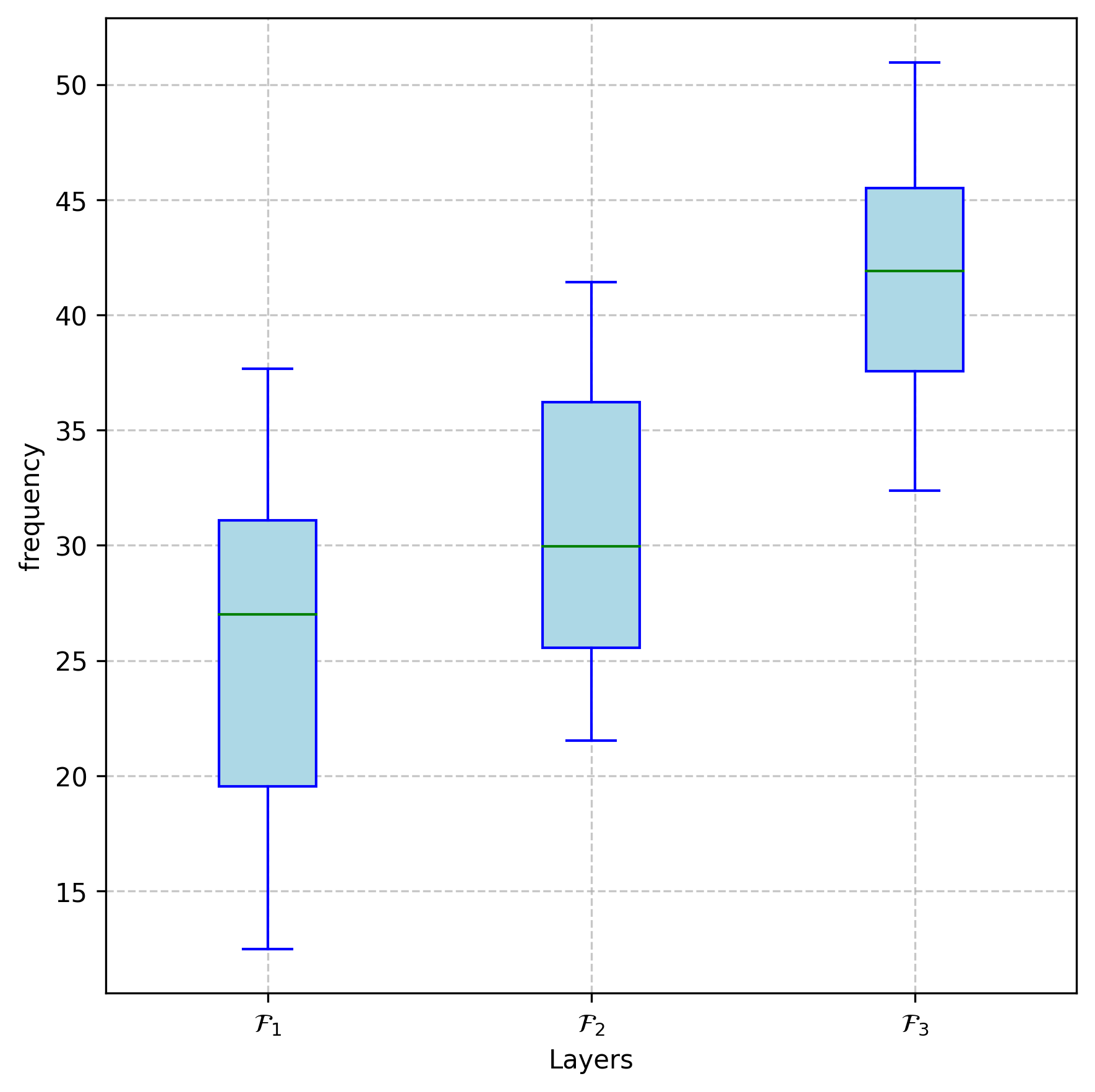}
  \caption{Visualization of the frequency distribution trends of activation functions across MLPs.}
  \label{fig:box}
\end{minipage}
\begin{minipage}[c]{0.35\textwidth}
  \centering
  \includegraphics[width=\linewidth]{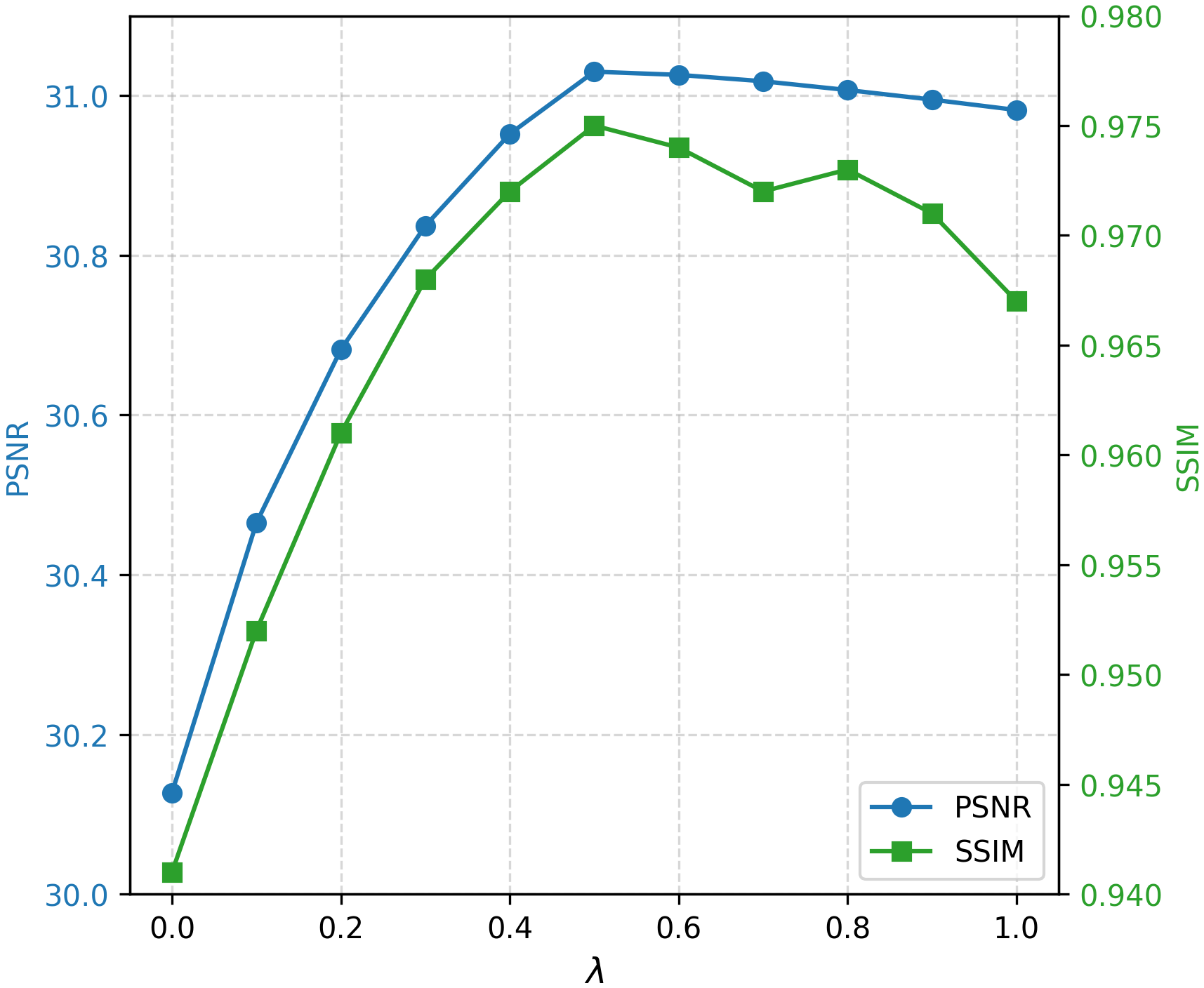}
  \caption{Ablation study between different choices of $\lambda$.}
  \label{fig:ab_lambda}
\end{minipage}
\begin{minipage}[c]{0.35\textwidth}
  \centering
  \includegraphics[width=\linewidth]{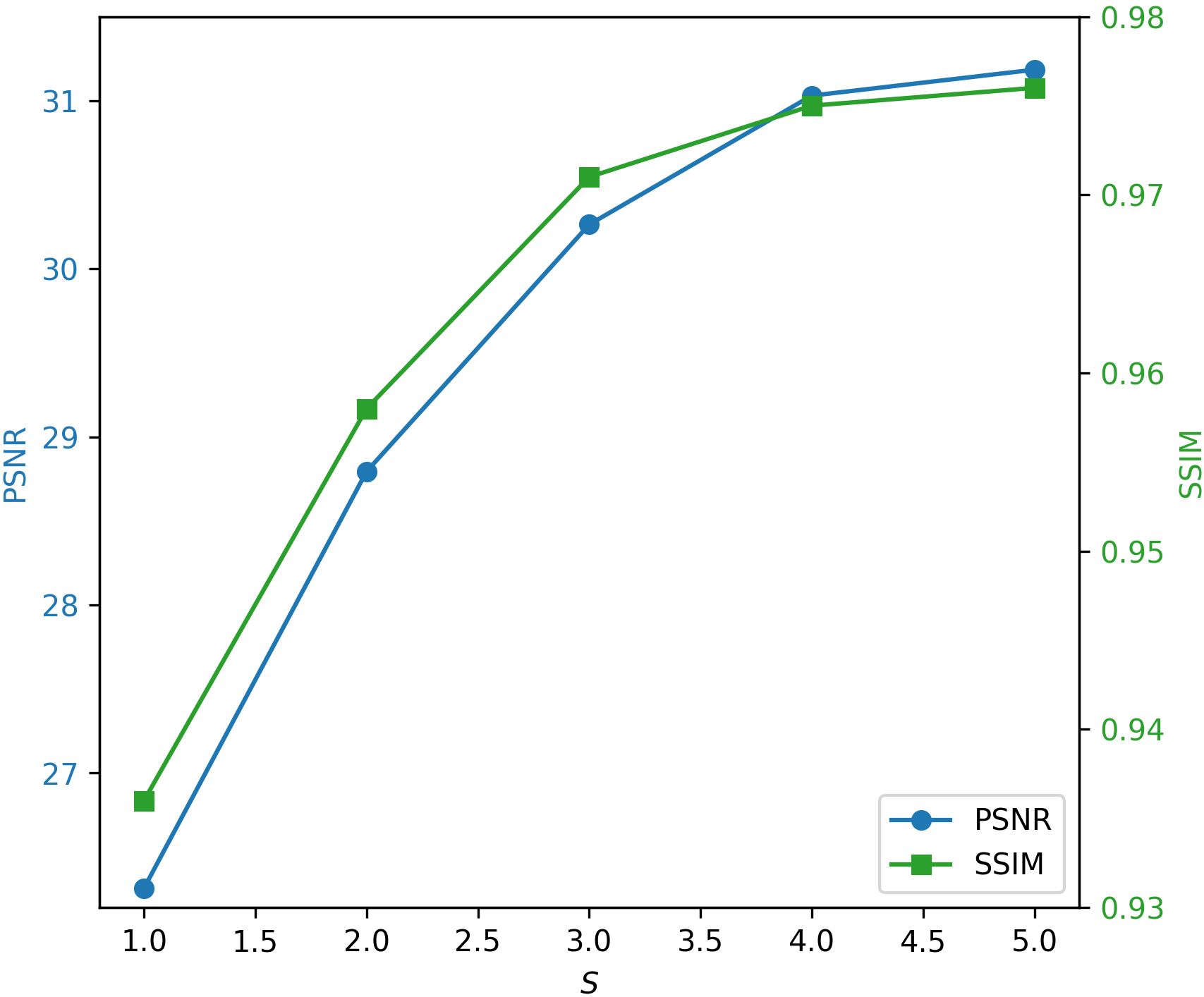}
  \caption{Ablation study for different choices of $S$.}
  \label{fig:ab_S}
\end{minipage}

\end{figure*}

\textbf{Qualitative Results.} Figure~\ref{fig:results} presents visualization results of the reconstructed human motion sequences generated by our method under $\times 2$, $\times 3$, $\times 4$, and $\times 5$ scales. The results demonstrate that our approach not only preserves the global structural integrity of the motions but also effectively captures subtle physical dynamics, such as velocity and acceleration. This leads to temporally smoother and more physically plausible high-FPS motion reconstructions.

To further evaluate the effectiveness of the proposed parametric activation function, we visualize the frequency variation trends of the learned activation functions, as presented in Figure~\ref{fig:box}. We visualize the top 50\% of frequencies corresponding to the largest absolute activation magnitudes. We can see that the model exhibits increasing sensitivity to higher-frequency components as the temporal scale decreases. This frequency-dependent behavior across different scales demonstrates the rationality and effectiveness of our proposed activation function.


\subsection{Motion Inbetweening }

\paragraph{Evaluation Metrics.}
Following previous work~\cite{oreshkin2023motion,qin2022motion}, we consider L2Q (the global quaternion squared loss), L2P (the global position squared loss) and NPSS (the normalized power spectrum similarity score). 
The L2P and L2Q measure the average L2 distance of the global joint position and rotation
(in quaternion) per joint per frame and NPSS evaluates angular differences between predicted motion and ground truth in the frequency domain. 

\textbf{Quantitative Results.} We compare our approach with MC-Trans~\cite{duan2021single}, NeMF~\cite{he2022nemf}, ERD-QV~\cite{harvey2020robust}, $\Delta$-Interp~\cite{oreshkin2023motion}, and TS-Former~\cite{qin2022motion} on the LAFAN1 dataset, with the experimental results presented in Table~\ref{fig:box}. Compared to existing methods, our model achieves significantly improved performance, highlighting both the effectiveness and the strong generalization capability of our approach.

\begin{figure*}[t]
   \centering
\includegraphics[width = 0.9\textwidth]{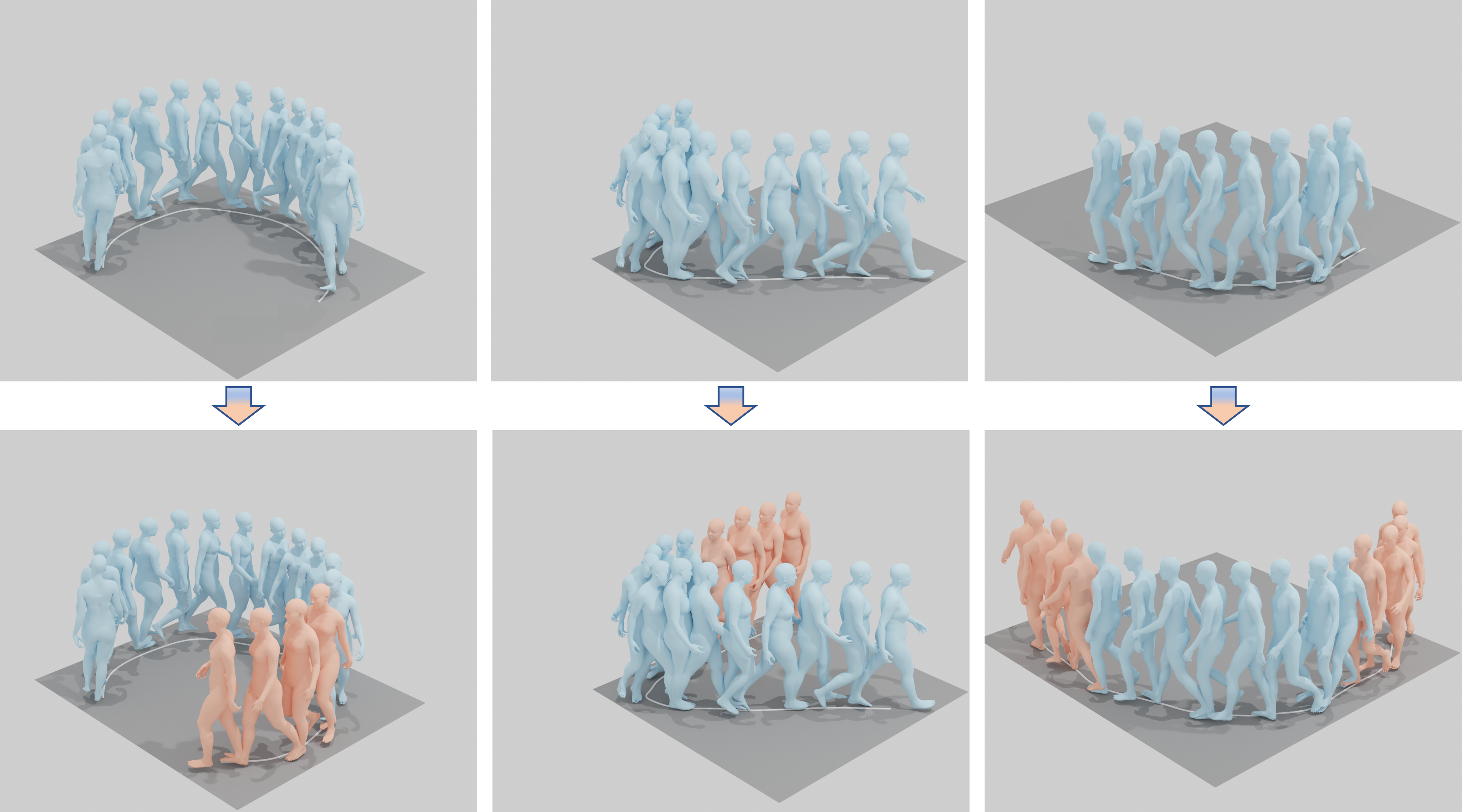}
  \caption{Examples of temporal inpainting Application. Blue motion indicates the range where the motion content is given by the reference sequence. Red motion indicates the range of motion content generated by our method.}
   \label{fig:app}
\end{figure*}

\subsection{Ablation Studies}

\textbf{The Ablation for PAID.} 
In our PAID module, we conducted an ablation study on the HumanML3D dataset under the $\times$4 scale setting to investigate two factors: the choice of $\zeta$ and whether the activation functions across different layers share parameters. As shown in Table~\ref{tab:zeta}, performance degrades when parameters are shared, indicating that applying distinct activations at different scales can enhance model performance. Additionally, we observe that increasing $\zeta$ consistently improves the results, though the performance gain diminishes significantly once $\zeta$ exceeds 16. To balance effectiveness and computational cost, we set $\zeta$=16 as the hyperparameter in our experiments.

\textbf{The choice of $\lambda$.} 
We conducted an ablation study under the 4 scale setting on the HumanML3D dataset to investigate the impact of the weighting factor $\lambda$ in the loss function. As shown in Figure~\ref{fig:ab_lambda}, incorporating velocity supervision significantly improves the model's performance. While an excessively large $\lambda$ leads to a slight performance drop, the overall results remain strong. Therefore, we set $\lambda$=0.5 as the hyperparameter in our experiments to balance accuracy and stability.

\textbf{The choice of $S$.} 
We performed an ablation study on the choice of the number of scales 
$S$ in our architectural design under the 4 scale setting on the HumanML3D dataset. As illustrated in Figure~\ref{fig:ab_S}, increasing the number of scales significantly improves model performance, with a substantial gain observed when increasing $S$ from 1 to 2. However, the performance improvement becomes marginal when $S$ exceeds 4. To strike a balance between effectiveness and computational cost, we set $S$=4 as the hyperparameter in our experiments.

\subsection{Application}
Leveraging the inherent properties of implicit neural representations (INRs) and the continuity of parametrically activated MLPs, our method is capable of generating plausible outputs for previously unseen temporal inputs $t$ during inference. Consequently, it supports both forward and backward extrapolation of motion sequences. Figure~\ref{fig:app} illustrates the results of our method. It can be observed that our method is capable of predicting motion either before, after, or simultaneously on both sides of the given input motion, and still achieves satisfactory results. This demonstrates the extendibility and flexibility of our approach.

\section{Conclusion}

In this paper, we presented \textbf{\textit{\NAME}}, a novel parametric activation-induced hierarchical implicit representation framework tailored for continuous modeling of human motion sequences across arbitrary frame rates. Unlike conventional implicit neural representation (INR) approaches designed primarily for image and video domains, \textbf{\textit{\NAME}} is specifically crafted to address the unique temporal and physical characteristics inherent in motion data. By leveraging a hierarchy of temporal encoders to extract multi-scale dynamic features and integrating a parametric activation function to enhance the representational capacity of the decoder, our method achieves high-fidelity reconstruction of complex motion patterns. Extensive experiments on several benchmark datasets and several settings demonstrate the effectiveness and generalizability of our approach, setting a new direction for high-quality, continuous human motion representation.
{
    \small
    \bibliographystyle{ieeenat_fullname}
    \bibliography{main}
}


\end{document}